\documentclass[runningheads]{llncs}
\usepackage{graphicx}
\usepackage{floatrow} 
\newfloatcommand{capbtabbox}{table}[][\FBwidth]   
\usepackage{amsmath}
\usepackage{algorithm}

\usepackage[noend]{algpseudocode}
\usepackage{color}
\usepackage{multirow}
\usepackage{bm}
\usepackage{pifont}
\newcommand{\cmark}{\ding{51}}%
\newcommand{\xmark}{\ding{55}}%
\begin{document}

\title{Self-Referenced Deep Learning} 
\titlerunning{Self-Referenced Deep Learning} 


\author{Xu Lan\inst{1} 
	\and
	Xiatian Zhu\inst{2} 
	\and
	Shaogang Gong\inst{1} 
}
\institute{$^1$ Queen Mary University of London,\\
	\email{x.lan@qmul.ac.uk}, \email{s.gong@qmul.ac.uk}	\\
	$^2$ Vision Semantics Ltd \\
	\email{eddy@visionsemantics.com}
}

\authorrunning{Xu Lan et al.} 



\maketitle


\begin{abstract}
	Knowledge distillation is an effective approach to 
	transferring knowledge from a teacher neural network to a student target network
	for satisfying the low-memory and fast running requirements in practice use.
	Whilst being able to create stronger target networks compared to
	the vanilla non-teacher based learning strategy,
	this scheme needs to train {\em additionally} a large teacher model
	with expensive computational cost. In this work, we present a Self-Referenced Deep Learning (SRDL)
	strategy. Unlike both vanilla optimisation and existing knowledge distillation,
	SRDL distils the knowledge discovered by
	the in-training target model {\em back} to itself
	to regularise the subsequent learning procedure {\color{black} therefore eliminating 
		the need for training a large teacher model.}
	SRDL improves the model generalisation performance 
	compared to vanilla learning and 
	conventional knowledge distillation approaches 
	with {\em negligible} extra computational cost.
	Extensive evaluations show that a variety of deep networks 
	benefit from SRDL resulting in enhanced deployment performance
	on both coarse-grained object categorisation tasks (CIFAR10, CIFAR100, Tiny ImageNet, {and ImageNet})
	and fine-grained person instance identification tasks (Market-1501).
\end{abstract}

\section{Introduction}

Deep neural networks have been shown to be effective for solving many
computer vision tasks \cite{krizhevsky2012imagenet,simonyan2014very,szegedy2015going,he2016deep,lan2017deep,li2018harmonious}.
However, they are often computationally expensive 
due to having very deep and/or wide architectures
with millions of parameters \cite{zagoruyko2016wide,he2016deep,simonyan2014very}.
This leads to slow execution and the need for large storage,
reducing their deployability to situations with low
memory and  limited computing budget, e.g. mobile phones.
This has given rise to efforts in developing more compact models, such as
parameter binarisation \cite{rastegari2016xnor},
filter pruning \cite{li2016pruning},
model compression \cite{han2015deep},
and knowledge distillation \cite{hinton2015distilling}.

\begin{figure}
	\centering
	\includegraphics[width=1.0\linewidth]{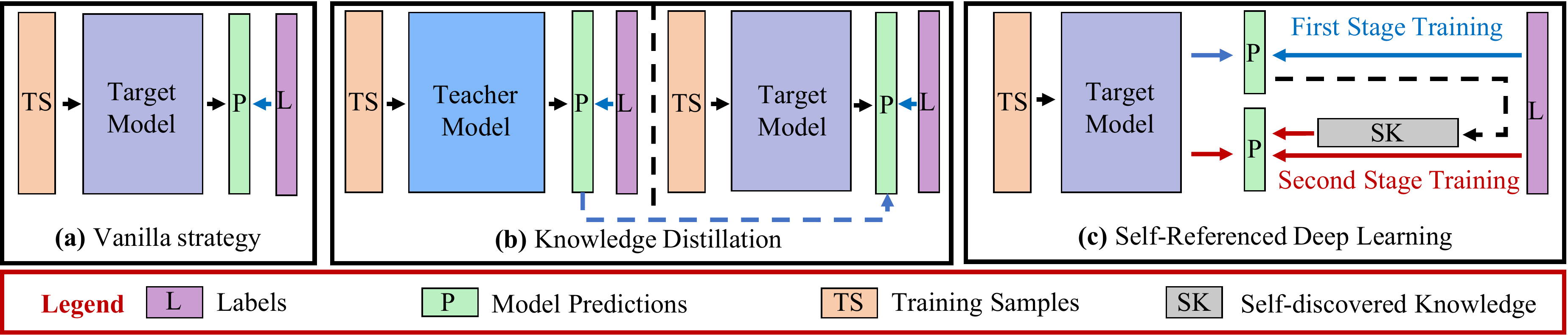} 	
	\caption{ 
		Illustration of three  different deep network learning methods.
		{\bf(a)} The vanilla  training: 
		Optimise the target model from the supervision of training label 
		for $M$  epochs in one stage.
		{\bf(b)} The Knowledge Distillation training:
		Firstly learn a teacher model in a {\em computationally intensive} manner;
		Then extract the learned knowledge  from the teacher model;
		Lastly optimise the target model by leveraging both the label data and the teacher's knowledge for $M$ epochs.
		{\bf(c)} The proposed SRDL training:
		In the first stage, learn the target model by the label supervision
		for half ($M/2$) epochs (similar to (a) but with a different learning rate strategy), and extract intermediate knowledge from the thus-far trained model (similar to (b) but without a heavy teacher model to train);
		In the second stage, continuously optimise the target model 
		from the joint supervision of the labelled training data and the self-discovered knowledge for another half $(M/2)$ epochs.}
	
	\label{fig:stagetraining}
\end{figure}

Among these existing techniques, knowledge distillation
\cite{hinton2015distilling} is a generic approach suitable to a wide variety of networks and applications.
It is based on the observation
that compared to large networks, small networks often have similar representation capacities 
but are harder to define and train the parameters of a target function
\cite{ba2014deep,bucilu2006model}. 
As a solution to this challenge, knowledge distillation first trains 
a deeper and/or wider ``teacher'' network (or an ensemble model), 
then learns a smaller ``student'' network to imitate the teacher's
classification probabilities \cite{hinton2015distilling} and/or feature
representations \cite{ba2014deep,romero2014fitnets} 
(Fig \ref{fig:stagetraining}(b)). 
This imposes additional information
beyond conventional supervised learning signals (Fig \ref{fig:stagetraining}(a)),
leading to a more discriminative student model than learning the target model 
without the teacher's knowledge. 
%
However, this generalisation improvement comes with significant extra computational cost and
model training time of the teacher model.

{\color{black}
	In contrast to knowledge distillation, fast model optimisation aims to reduce the cost of 
	training a target model. While it is relatively fast to train a model on small datasets such as CIFAR10 \cite{krizhevsky2009learning} in a few hours, 
	model training on larger datasets like the ILSVRC dataset \cite{russakovsky2015imagenet} 
	requires a few weeks.
	Hence, fast optimisation in deep model training has increasingly become an important problem to be addressed.
	There are several different approaches
	to fast optimisation,  such as 
	model initialisation \cite{erhan2010does,mishkin2015all} and
	learning rate optimisation \cite{kingma2014adam,zeiler2012adadelta,duchi2011adaptive}}.
%

In this work, we aim to jointly solve both knowledge 
distillation for model compression and fast optimisation in model
learning using a unified deep learning strategy. 
%
To that end, we propose a {\bf Self-Referenced Deep Learning} (SRDL) strategy that
integrates the knowledge distillation concept
into a vanilla network learning procedure (Fig \ref{fig:stagetraining}(c)).
%
Compared to knowledge distillation, SRDL exploits different and 
{\em available} knowledge without the 
need for additionally training an expensive teacher by
self-discovering knowledge with the target model itself during
training.
%
Specifically, SRDL begins with 
training the target network by a conventional supervised learning objective
as a vanilla strategy,
then extracts self-discovered knowledge (inter-class correlations) 
during model training,
and continuously trains the model until convergence by satisfying two losses concurrently:
a conventional supervised learning loss, and 
an {\em imitation loss} that regulates the classification probability
predicted by the current (thus-far) model with the self-discovered knowledge.
By doing so, the network learns significantly better than
learning from a conventional supervised learning objective alone, as
we will show in the experiments.

Our {\bf contributions} are:
{\bf (I)} 
We investigate for the first time the problems of knowledge 
distillation based model compression 
and fast optimisation in model training using a unified deep learning approach, 
an under-studied problem although both problems
have been studied independently in the literature.
{\color{black}{\bf (II)}
	We present a stage-complete learning rate decay schedule
	in order to maximise the quality of intermediate self-discovered knowledge 
	and therefore  avoid the negative guidance to the
	subsequent second-stage model optimisation.
	%
	{\bf (III)}
	We further introduce a random model restart scheme
	for the second-stage training with the purpose
	of breaking the optimisation search space constraints
	tied to the self-referenced deep learning process.}

Extensive comparative experiments are conducted
on object categorisation tasks (CIFAR10/100 \cite{krizhevsky2009learning},
Tiny ImageNet \cite{le2015tiny}, and ImageNet \cite{russakovsky2015imagenet})
and person instance identification tasks (Market-1501 \cite{zheng2015scalable}).
These results show that the proposed SRDL offers a favourable trade-off
between model generalisation and model capacity (complexity).
It narrows down the model performance gap between 
the vanilla learning strategy and
knowledge distillation with almost no extra computational cost.
In some cases, SRDL even surpasses the performance of
conventional knowledge distillation whilst maintaining the model learning
efficiency advantage.

\section{Related Work}

\noindent {\bf Knowledge Distillation. }
Model compression by knowledge distillation was firstly studied in
\cite{bucilua2006model} and recently re-popularised by
Hinton et al. \cite{hinton2015distilling}.
The rationale behind distillation
is the introduction of extra supervision from a teacher model in training the target model
in addition to a conventional supervised learning objective such as the
cross-entropy loss subject to the labelled training data.
The extra supervision were typically obtained from a pre-trained powerful teacher model
in the form of classification probabilities \cite{hinton2015distilling},
feature representation \cite{ba2014deep,romero2014fitnets},
or inter-layer flow (the inner product of feature maps) \cite{yim2017gift}.
Knowledge distillation has been exploited to distil easy-to-train large networks 
into harder-to-train small networks \cite{romero2014fitnets},
or transfer high-level semantics to earlier layers \cite{lan2018person},
or simultaneously enhance and transfer knowledge on-the-fly \cite{lan2018knowledge}.
%

Recently, some theoretical analysis have been provided to relate distillation to information learning theory
for which a teacher provides privileged information
(e.g. sample explanation) to a student in order to facilitate fast learning
\cite{lopez2015unifying,vapnik2015learning}.
Zhang et al. \cite{zhang2016real} exploited this idea for video based action recognition
by considering the computationally expensive optic flow as privileged information 
to enhance the learning of a less discriminative motion vector model.
This avoids the high cost of computing optic flow in model deployment
whilst computing cheaper motion vectors enables real-time performance.

In contrast to all the above existing works, we aim to eliminate the extra teacher model training all together.
To this end, we uniquely explore self-discovered knowledge in target model
training by {\em self-distillation}, therefore more cost-effective. Concurrent with our work, Furlanello et al.
\cite{furlanello2018born} independently proposes training
the networks in generations, in which the next generation is jointly guided by the standard one-hot classification labels and the knowledge learned in the previous generation. However, the training budget of each generation is almost the same as the vanilla strategy, 
leading to the total cost of this method several times 
more expensive than vanilla training.

\noindent {\bf Fast Optimisation. }
Fast optimisation of deep neural networks has gained increasing attention
for reducing the long model training time by rapid model learning convergence \cite{krahenbuhl2015data,saxe2013exact}.
A simple approach is by Gaussian initialisation with zero mean and unit variance,
and Xavier initialisation \cite{glorot2010understanding}.
But these are not scalable to very deep networks. More recent
alternatives have emerged \cite{he2015delving}.
The rational is that
good model initialisation facilitates model learning
to rapidly reach the global optimum with minimal vanishing and/or exploding
gradients. 
Additional options include improved optimisation algorithms to 
mitigate the slow convergence of SGD
by 
sidestepping saddle points in the loss function surface \cite{johnson2013accelerating},
and learning rate refinement 
that exploits a cycle rate to train a neural network in 
the context of ensembling multiple models \cite{huang2017snapshot}.
However, model ensemble multiplies the deployment cost by times.

In the spirit of fast model optimisation, our method 
aims to achieve more generalisable model
learning without extra computational cost for learning an independent teacher model.
By self-distillation, the proposed method can improve the performance of
both small and large networks, so it is generally applicable.

\section{Self-Referenced Deep  Training}

\subsection{Problem Statement}
For supervised model learning,
we assume $n$ labelled training samples $\mathcal{D}=\{(\bm{I}_{i}, {y}_{i})\}_{i}^{n}$. 
Each sample belongs to one of $C$ classes 
${y}_{i} \in \mathcal{Y} = [1,2,\cdots,C]$,
with the ground-truth label typically represented as a one-hot vector.
The objective is to learn a classification deep CNN model
generalisable to unseen test data
through a cost-effective training process. 

{\color{black}
	In this work, we formulate a novel deep learning approach
	that improves the model generalisation capability
	through employing self-discovered knowledge as additional supervision signal
	with marginal extra computational cost and hence 
	not hurting the computing scalability.
	We call this strategy {\bf Self-Referenced Deep Learning} (SRDL).
	We begin with revisiting the vanilla deep model training method (Fig \ref{fig:stagetraining}(a)) 
	before elaborating the proposed SRDL approach.}%

{\color{black}\subsection{Vanilla Deep Model Training}
	\label{sec:CNN}
	For training a classification deep model, 
	the softmax cross-entropy loss function is usually adopted. 
	Specifically, we predict the posterior probability of a labelled sample $\bm{I}$
	over any class ${c}$ via the softmax criterion:
	\begin{equation}
	{p}(c | \bm{x}, \bm{\theta}) = \frac{\exp(z_c)} {\sum_{j=1}^{C} \exp(z_j)}, \;\; 
	z_j = \bm{W}_{j}^{\top} \bm{x}, \;\; c \in \mathcal{Y}
	\label{eq:softmax}
	\end{equation}
	where
	$\bm{x}$ refers to the embedded feature vector of $\bm{I}$,
	$\bm{W}_j$ the $j$-th class prediction function parameter, and
	$\bm{\theta}$ the neural network model parameters.
	We then compute the cross-entropy loss on a labelled sample $\bm{x}$ (in a mini-batch) as:
	\begin{equation}
	\mathcal{L}_\text{ce} = 
	\log \Big(p({y}|\bm{x}) \Big)
	\label{eq:loss}
	\end{equation}
	where $y$ specifies the ground-truth label class of $\bm{x}$.

	\noindent{\bf Discussion. }
	For a model subject to the vanilla training (Fig \ref{fig:stagetraining}(a)),
	the cross-entropy loss is utilised to supervise the 
	model parameters (e.g. by the stochastic gradient descent algorithm)
	iteratively in a {\em one-stage} procedure.
	%
	This training method relies {\em only} on the supervision of 
	per-sample label, but {\em ignores} the discriminative knowledge
	incrementally discovered by the in-training model itself.
	It may lead to sub-optimal optimisation. We overcome this problem by introducing a mechanism to exploit self-discovered intermediate knowledge in a computationally economic manner.
}

\subsection{Self-Referenced Deep Learning}
\label{sec:SRDL}

\begin{figure*} [h]
	\centering
	\includegraphics[width=1.0\linewidth]{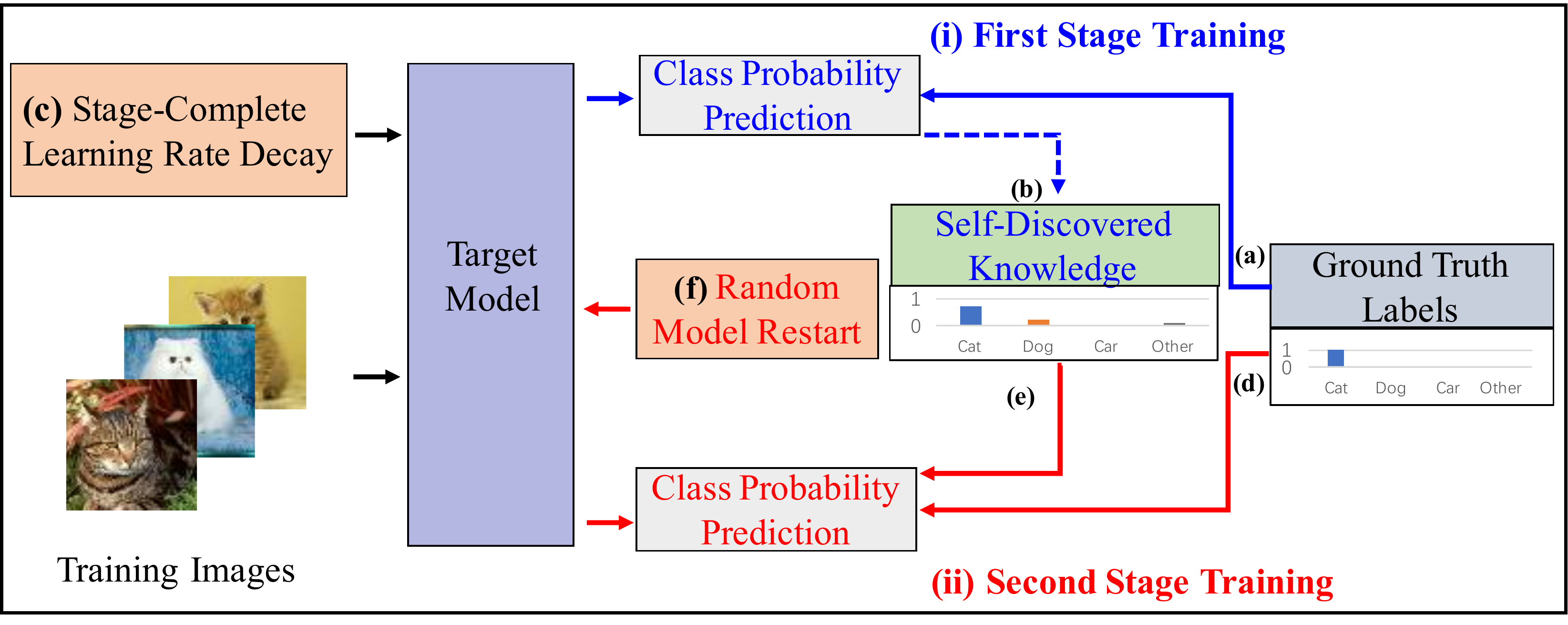} 	
	\caption{{\color{black}
			Overview of the proposed  Self-Referenced Deep Learning (SRDL). The SRDL strategy consists of two stages training:
			\textbf{First stage}: We train the target model by 
			a cross-entropy loss (Eq \eqref{eq:loss}) with {\bf(a)} the available
			label supervision for half epochs, whilst learning to {\bf(b)} extract discriminative intermediate knowledge 
			concurrently (Eq \eqref{eq:softmax_soft}). 
			To maximise the quality of self-discovered knowledge, we introduce {\bf(c)} a pass-complete
			learning rate decay schedule (Eq \eqref{eq:pass_decay}).
			\textbf{Second stage}: 
			we continuously optimise the target model
			for the other half epochs by the joint supervision (Eq \eqref{eq:Total_loss}) 
			of both {\bf(d)} the label data
			and {\bf(e)} self-discovered intermediate knowledge in an end-to-end manner.
			We {\bf(f)} randomly restart the model for the second stage
			to break the optimisation search space constraint from self-referenced deep learning mechanism.}}
	
	\label{fig:piplline}
	
\end{figure*}

\noindent{\bf SRDL Overview. }
The proposed SRDL approach is a knowledge referenced end-to-end deep model
training strategy. 
The overview of our SRDL approach is depicted in Fig \ref{fig:piplline}.
This is realised through reformulating the vanilla  training process
into two equal-sized stages: 
\begin{enumerate}
	\item In the first stage (Fig \ref{fig:piplline}(i)), 
	SRDL learns the target model as a vanilla algorithm with a
	conventional supervised learning objective,
	while tries to induce reliable knowledge.
	\item In the second stage (Fig \ref{fig:piplline}(ii)), 
	SRDL continues to train the model by
	a conventional supervised loss and 
	a self-discovered knowledge guided imitation loss
	concurrently.
\end{enumerate}

{
	For model training, 
	SRDL consumes the same number of epochs as the vanilla counterpart.
	The extra marginal cost is due to self-discovered knowledge extraction
	(see \texttt{Evaluation Metrics} in Sec \ref{sec:exp}).
	Consequently, SRDL allows to benefit model generalisation as knowledge distillation 
	at faster optimisation speed.
	Once the target model is trained, it is deployed to the test data 
	same as the vanilla method.}

\noindent{\bf (I) First Stage Learning. }
In the first stage of SRDL, 
we train the deep model $\bm{\theta}$ by the cross-entropy loss Eq \eqref{eq:loss}.
%
Model training is often guided by 
a learning rate decay schedule
such as the 
step-decay function \cite{he2016deep,huang2017densely}:
\begin{equation} 
\label{eq:decay}
\epsilon_t= \epsilon_0 \times f_\text{step}(t,M), \;\; t \in [1,\cdots, M]
\end{equation}
where $\epsilon_t$ denotes the learning rate at the $t$-th epoch 
(initialised as $\epsilon_0$, in total $M$ epochs),
and $f_\text{step}(t,M)$ the step-decay function.
The learning rate decay aims to encourage the model to converge to a satisfactory local minimum
without random oscillation in loss reduction during model training.
%
However, if applying the conventional step-decay scheme
throughout the optimisation process,
SRDL may result in premature knowledge during training.
%
This is because, the model still resides in an unstable local minimum
due to that the learning rate drop is not sufficiently quick
\cite{welling2011bayesian}.

\begin{figure}[h]
	\centering
	\includegraphics[width=0.9\linewidth]{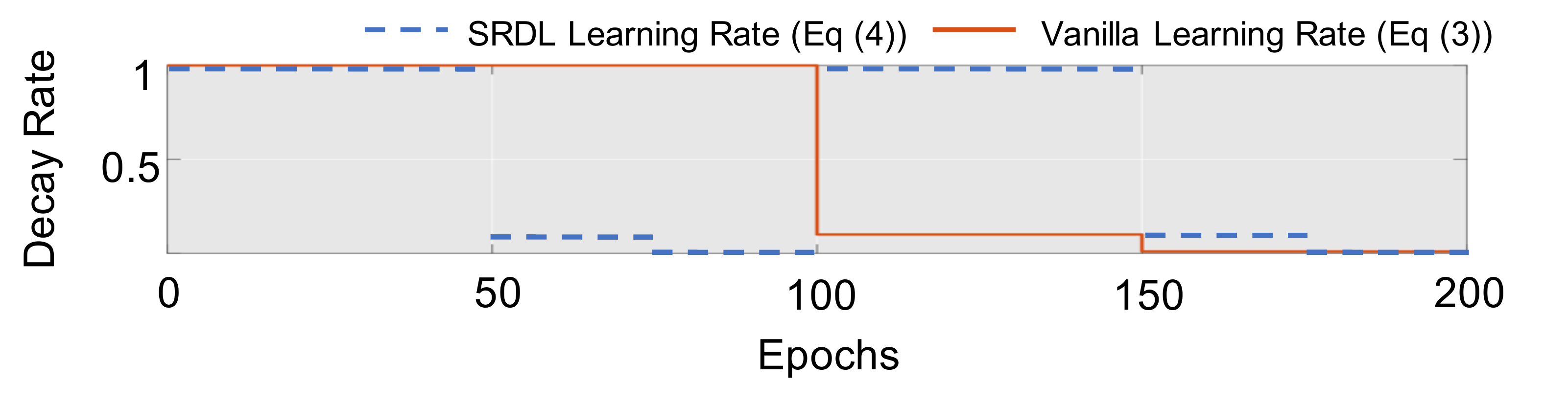} 	
	\caption{Illustration of a  vanilla learning rate step-decay function and 
		the proposed stage-complete learning rate step-decay schedule.}
	\label{fig:step_func}
\end{figure}

To overcome this problem, 
we propose to deploy an {\em individual} and {\em complete} step-decay schedule
for both first and second stages of SRDL (Fig \ref{fig:piplline}(c)), {\color{black}subject to the condition of remaining the same training epochs (cost).} 
Formally, this schedule is expressed as:
\begin{equation} 
\label{eq:pass_decay}
\epsilon_t= \epsilon_0 \times f_\text{step}(t,0.5M)
\end{equation}
The intuition is that, the in-training model can be {\em temporarily} 
pushed towards a reasonably stable local minimum
within the same number of (e.g. 0.5$M$) epochs to 
achieve a more-ready state therefore help ensure
the quality of self-discovered knowledge.
We call this a \textbf{\em stage-complete} learning rate step-decay schedule (Fig \ref{fig:step_func}).
{\color{black}
	Our evaluations verify the significance of this design 
	while guaranteeing the goodness of the self-referenced knowledge 
	(see Table \ref{tab:LR_schedule}).}

At the end of the first stage of SRDL with 
a ``half-trained'' model (denoted as $\bm{\theta}^{*}$), 
we extract the self-discovered knowledge in the form of 
per-sample class probability prediction (Fig \ref{fig:piplline}(b)).
Formally, we compute the class probability
for each training sample $\bm{x}$ by
a {\em softened} softmax operation as:
\begin{equation}
\label{eq:softmax_soft}
\tilde{p}(c | \bm{x}, \bm{\theta^}{*}) = \frac{\exp(z_c/T)} {\sum_{j=1}^{C} \exp(z_j/T)}, \;\; 
z_j = \bm{W}_{j}^{\top} \bm{x}, \;\; c \in \mathcal{Y}
\end{equation}
where the temperature parameter $T$ controls the
softening degree, with higher values meaning more softened predictions.
We set $T\!\!=\!\!3$ in our experiments as suggested in \cite{hinton2015distilling}.

\noindent{\bf (II) Second Stage Learning. }
To improve the generalisation performance of the model,
we use the self-discovered knowledge to provide training experience
at second stage model learning in SRDL.
We quantify the imitation of the current model to
the knowledge $\tilde{p}(j|\bm{x}, \bm{\theta}^*)$ with
Kullback Leibler (KL) divergence (Fig \ref{fig:piplline}(e)), formulated as: 
\begin{equation} 
\label{eq:kl_loss}
R_\text{kl}= \sum_{j=1}^{C}   { \tilde{p}(j|\bm{x}, \bm{\theta}^*)} \log \frac {\tilde{p}(j|\bm{x}, \bm{\theta}^*)}{{\tilde{p}(j|\bm{x}, \bm{\theta})}}.
\end{equation}
where $\tilde{p}(j|\bm{x}, \bm{\theta})$ is the class probability prediction of the up-to-date 
model $\bm{\theta}$ computed by Eq \eqref{eq:softmax_soft}.
The overall loss function for the second stage in SRDL is:
\begin{equation} 
\label{eq:Total_loss}
\mathcal{L}= \mathcal{L}_\text{ce} + T^2 * R_\text{kl}
\end{equation}
with the squared softening-temperature (Eq \eqref{eq:softmax_soft})
as the balance weight.
{\color{black}
	The gradient magnitudes produced by the soft targets $\tilde{p}$ 
	are scaled by $\frac{1}{T^2}$, 
	so we multiply the distillation loss term by
	a factor ${T^2}$ to ensure that 
	the relative contributions of ground-truth and teacher probability distributions 
	remains.
}
In doing so, the network model learns to both
predict the correct class label (cross-entropy loss $\mathcal{L}_\text{ce}$) 
and 
align the class probability of previous training experience (imitation loss $R_\text{kl}$)
concurrently.

\noindent \textbf{\em Random Model Restart. } 
A key difference between SRDL and knowledge distillation
is that SRDL enables a model to learn from its own (previously
revealed) knowledge through training experience rather than from an independent teacher's knowledge.
This self-discovered knowledge is represented in the ``half-trained'' model parameters $\bm{\theta}^*$. 
If we further train the model at the second stage from $\bm{\theta}^*$
by Eq \eqref{eq:Total_loss}, the learning may become less explorable for
better local or global minimum 
due to the stacking effect of the imitation loss and the model parameter status. 
Therefore, 
we start the second stage training 
with {\em randomly} initialised model parameters.

This scheme is based on three considerations:
(1) A large proportion of the knowledge learned in the first stage has been extracted 
and used in the second stage.
(2) The same training data will be used.
(3) Random initialisation offers another opportunity for the model
to converge to a better local minimum.
%
Our experiment validates the effectiveness of this {\em random restart} scheme (see Table \ref{tab:restart} in Sec \ref{sec:exp_analysis}).
%

SRDL model training is summarised in Alg  \ref{alg}.
In our experiments, a SRDL trained model is tested against both the vanilla model training
strategy and the knowledge distillation method.

\begin{algorithm}
	\caption{Self-Referenced Deep Learning}\label{alg}
	\begin{algorithmic}[1]
		\State{{\bf Input}: Labelled training data $\mathcal{D}$; Training epochs $M$;
		}
		\State{{\bf Output}: Trained CNN model $\bm{\theta}$;}

		\State{{\bf (I) First stage learning} }
		\State{{\bf Initialisation}: t=1; Random model $\bm{\theta}$ initialisation;}
		\While{$t \leq 0.5*M$}
		\State (i) Update the learning rate $\epsilon_t$  (Eq \eqref{eq:pass_decay});
		\State (ii) Update $\bm{\theta}$ by cross-entropy loss (Eq \eqref{eq:loss});
		\EndWhile
		\State \textbf{end}

		\State{{\bf Knowledge Extraction}
			Induce per-sample class probability predictions (Eq \eqref{eq:softmax_soft}); }

		\State{{\bf (II) Second stage learning} }
		\State{{\bf Initialisation}: t=1; Random model $\bm{\theta}$ restart;}
		\While{$t \leq 0.5*M$}
		\State (i) Update the learning rate $\epsilon_t$  (Eq \eqref{eq:pass_decay});
		\State (ii) Update $\bm{\theta}$ by soft-feedback referenced loss (Eq \eqref{eq:Total_loss});
		\EndWhile
		\State \textbf{end}	
	\end{algorithmic}
\end{algorithm}

\section{Experiments}
\label{sec:exp}

\subsection{Experimental Setup}

\noindent {\bf Datasets. }
For experimental evaluations, we use four benchmarking datasets
including both coarse-grained object classification
and fine-grained person instance identification 
Specifically, the \textbf{\em CIFAR$10$} and
\textbf{\em CIFAR$100$} \cite{krizhevsky2009learning}
datasets contain $32\!\times\!32$ sized natural images
from 10 and 100 object classes.
Both adopt a 50,000/10,000 train/test image split.
The \textbf{\em Tiny ImageNet} \cite{le2015tiny} consists of 110,000 
64$\!\times\!$64 images from 200 object classes.
We adopt the standard 100,000/ 1,000 train/val setting.
%
The \textbf{\em ImageNet \cite{krizhevsky2012imagenet}} is a large scale 1,000-class object image classification benchmark, 
providing 1.2 million images for
training, and 50,000 images for validation.
The \textbf{\em Market-1501} \cite{zheng2015scalable} is a
person re-identification dataset.
Different from image classification as tested in the above four datasets, 
person re-identification is a more fine-grained recognition problem
of matching person instance across non-overlapping camera views.
It is a more challenging task due to the inherent
zero-shot learning knowledge transfer from seen classes (identities)
to unseen classes in deployments, i.e. 
no overlap between training and test classes.
Market-1501 has 32,668 images of 1,501 different identities (ID)
captured by six outdoor cameras.
We use the standard 751/750 train/test ID split.
Following \cite{sun2017svdnet,li2017person},
we train the network by the cross-entropy loss (Eq \eqref{eq:loss})
and use the feature layer's output
as the representation of person bounding box images for test
by the Euclidean distance metric.
%

%

%

\noindent {\bf Performance Metrics. }
For performance measurement, we adopt the top-1 classification accuracy for image classification,
the standard Cumulative Matching Characteristic (CMC) accuracy (Rank-$n$ rates)
and mean Average Precision (mAP) for person instance recognition (re-id).
The CMC is computed for each individual rank $k$ as
the cumulative percentage of the truth matches for probes returned
at ranks $\leq k$.
And the Rank-1 rate is often considered as the most important performance indicator
of an algorithm's efficacy.
The mAP is to measure the recall of multiple
truth matches, computed by first computing the area under
the Precision-Recall curve for each probe, then calculating
the mean of Average Precision over all probes.
We measure the model optimisation complexity
with the {FLoating-point OPerations} (FLOPs):
\texttt{\small Forward-FLOPs} * \texttt{\small Epochs} * \texttt{\small Training-Set-Size}.

\noindent {\bf Neural Networks. }
We use 7 networks in our experiments:
one typical student net, ResNet-32 \cite{he2016deep};
two typical teacher nets, ResNet-110 \cite{he2016deep} and Wide ResNet WRN-28-10 \cite{zagoruyko2016wide};
and four varying sized nets, ResNet-50, DenseNet-121, DenseNet-201 and DenseNet-BC ($L$=190, $k$=40) \cite{huang2017densely}.

%
%


\noindent {\bf Implementation Details. }
%
For all three image classification datasets, we use SGD with Nesterov momentum
and set the mini-batch size to 128,
the initial learning rate to 0.1, 
the weight decay to 0.0002, 
and the momentum to 0.9. 
For Market-1501,
we use the same SGD but
with the mini-batch size of 32.
We assign sufficient epochs to all models to ensure convergence. On CIFAR datasets, the training budget is 300 epochs for DenseNet, and 200 epochs for ResNet and Wide ResNet models, same as \cite{huang2017snapshot}. We set 150/120 epochs on Tiny ImageNet/Market-1501 for all models.
%
All model optimisation methods take the same epochs to train the target networks.
We adopt a common learning rate decay schedule \cite{huang2017snapshot}: 
the learning rate drops by 0.1 at the 
50\% and 75\% epochs.  
%
%
The data augmentation includes horizontal flipping and 
randomly cropping from images padded
by 4 pixels on each side with missing pixels filled by
original image reflections \cite{he2016deep}.
We report the average performance of 5 independent runs for each experiment.



\begin{table} [h]
	\setlength{\tabcolsep}{0.1cm}
	\begin{center}
		\begin{tabular}{l|c|c|c|c|c|c|c}
			\hline
			Dataset & \multirow{2}{*}{\# Param} & \multicolumn{2}{c|}{CIFAR10} & \multicolumn{2}{c|}{CIFAR100} & 
			\multicolumn{2}{c}{Tiny ImageNet} \\
			\cline{1-1}\cline{3-8}
			Metrics & & Acc & TrCost & Acc & TrCost & Acc & TrCost \\
			\hline 
			\hline
			ResNet-32+vanilla
			& \multirow{3}{*}{0.5M}
			& 92.53& 0.08
			& 69.02 & 0.08
			& 53.33 & 0.32  \\
			ResNet-32+{\bf SRDL}
			&
			&\bf 93.12 &0.08
			& \bf 71.63 &0.08
			& \bf 55.53 &0.32 \\
			\cline{1-1}\cline{3-8}
			\color{black} Gain (SRDL-vanilla)
			&  
			& \color{black} +0.59 & 0
			&\color{black} +2.61 & 0
			&\color{black} +2.20 & 0 \\
			\hline\hline
			WRN-$28$-$10$+vanilla
			& \multirow{3}{*}{36.5M}
			& 94.98 &12.62
			& 78.32&12.62
			&  58.38 &50.48 \\
			WRN-$28$-$10$+{\bf SRDL}
			&
			&\bf 95.41 &12.62
			&\bf79.38 &12.62
			& \bf60.80 &50.48\\	
			\cline{1-1}\cline{3-8}
			\color{black} Gain (SRDL-vanilla)
			&
			&\color{black} +0.43 & 0
			&\color{black} +1.06 & 0
			&\color{black} +2.42 & 0 \\
			\hline
			\hline
			DenseNet-BC+vanilla
			& \multirow{3}{*}{25.6M}
			&\bf\color{blue}96.68 &10.24
			&\bf\color{blue}82.83 &10.24
			&\bf\color{blue}62.88 &40.96 \\
			DenseNet-BC+\bf SRDL
			&
			&\bf\color{red}96.87 &10.24
			&\bf\color{red}83.59 &10.24
			&\bf\color{red}64.19 &40.96  \\
			\cline{1-1}\cline{3-8}
			\color{black} Gain (SRDL-vanilla)
			&
			&\color{black} +0.19 & 0
			&\color{black} +0.76 & 0
			&\color{black} +1.31 & 0 \\
			\hline
		\end{tabular}
	\end{center}
	
	\vskip -0.32cm
	\caption{
		Comparison between SRDL and the vanilla learning strategy on image classification.
		Metric: Accuracy (Acc) Rate (\%).
		``Gain'': the performance gain by SRDL over vanilla.
		TrCost: Model training cost in unit of 10$^{16}$ FLOPs, {\bf lower is better}.
		M: Million.
		The first/second best results are in {\bf {\color{red} red}/{\color{blue} blue}}.
	}
	\label{tab:img_cls}
\end{table}

\subsection{Comparison with the Vanilla Learning Strategy}
We compared the image classification performance between
SRDL and the vanilla optimisation strategy.
We make the following observations from Table \ref{tab:img_cls}: 
\begin{enumerate}
	\item All three networks ResNet-32, WRN-28-10, and DenseNet-BC improve the
	classification performance when trained by the proposed SRDL.
	For example, ResNet-32 achieves an accuracy gain of 0.59\% on CIFAR10,
	of 2.61\% on CIFAR100,
	and of 2.20\% on Tiny ImageNet.
	This suggests the applicability of SRDL to standard varying-capacity network architectures.
	
	\item SRDL achieves superior model generalisation performance 
	with nearly zero extra model training cost\footnote{The computational cost of knowledge extraction required by both SRDL and Knowledge Distillation \cite{hinton2015distilling} is marginal (less than $\!0.67\%$ model training cost) and hence omitted for analysis convenience.}.

	\item Smaller network (ResNet-32) with fewer parameters
	generally benefits more from SRDL in model generalisation performance, 
	making our method more attractive to resource-limited applications.
	Hence, our SRDL addresses the notorious hard-to-train problem in small networks to some degree
	\cite{ba2014deep}.


\end{enumerate}

\noindent{\bf Results on ImageNet. }
We test the large scale ImageNet with DenseNet201
and obtain the Top-1/5 rates
$77.20\%$/$94.57\%$ by the vanilla vs {77.72\%/94.89\%} by our SRDL. This suggests that SRDL generalises to large scale object classification settings.

\subsection{Comparison with Knowledge Distillation}
We compared our SRDL with the closely related Knowledge Distillation
(KD) method \cite{hinton2015distilling}.
With KD, we take ResNet-32 as the target model,
WRN-28-10 and ResNet-110 as the pre-trained teacher models
to produce the per-sample class probability targets (i.e. the teacher's knowledge)
for the student. 
%
From Table \ref{table:KD} we draw these observations: 
\begin{table}[H]
	\setlength{\tabcolsep}{0.05cm}
	\begin{center}
		\begin{tabular}{c|c|c||c|c|c|c|c|c}
			\hline
			\multirow{2}{*}{Target Net}&	\multirow{2}{*}{Method} &	\multirow{2}{*}{Teacher Net} & \multicolumn{2}{c|}{CIFAR10} 
			& \multicolumn{2}{c|}{CIFAR100} & \multicolumn{2}{c}{Tiny ImageNet} \\
			\cline{4-9}
			&&& Acc & TrCost & Acc & TrCost & Acc & TrCost \\
			\hline\hline			
			\multirow{4}{*}{ResNet-32}
			& Vanilla
			& N/A
			& 92.53 &\color{red}\bf 0.08
			& 69.02 &\color{red}\bf 0.08
			& 53.33 &\color{red}\bf 0.32 
			\\
			\cline{2-9}
			&\multirow{2}{*}{KD}
			&WRN-28-10 (36.5M)&\color{blue}\bf 92.83& 12.70
			&\color{red} \bf 72.58 & 12.70
			&\color{red} \bf 56.80 & 50.80 
			\\ 	\cline{3-9}
			\bf & 
			&ResNet-110 (1.7M)& 92.75 & 0.30
			& 71.17 &  0.30
			& 55.06  & 1.20 
			\\  \cline{2-9}
			(0.5M) & \bf SRDL
			&N/A
			&\color{red} \bf 93.12 &\color{red} \bf 0.08 
			&\color{blue}\bf 71.63 &\color{red} \bf 0.08
			&\color{blue}\bf 55.53 &\color{red} \bf 0.32 
			\\
			\hline
		\end{tabular}
	\end{center}
	\vskip -0.32cm
	\caption{
		Comparison between SRDL and Knowledge Distillation (KD) on image classification. 
		Metric: Accuracy (Acc) Rate (\%).
		TrCost: Model training cost in unit of 10$^{16}$ FLOPs, {\bf lower is better}.
		Number in bracket: model parameter size.
		The first/second best results are in {\bf {\color{red} red}/{\color{blue} blue}}.
	}
	\label{table:KD}
	
\end{table}

\begin{enumerate}
	
	\item KD is indeed effective to improve small model generalisation
	compared to the vanilla optimisation,
	particularly when using a more powerful teacher (WRN-28-10).
	However, this is at the price of extra 157$\!\times$ (12.70/0.08-1 or
	50.80/0.32-1) model training cost.
	When using ResNet-110 as the teacher in KD, the performance gain is less significant.
	
	\item SRDL approaches the performance of KD(WRN-28-10) on CIFAR100 and Tiny ImageNet,
	whilst surpasses it on CIFAR10. 
	This implies that while small model is inferior to KD in self-discovering
	knowledge among a large number of classes,
	it seems to be superior for small scale tasks with fewer classes.
	
	\item SRDL consistently outperforms KD(ResNet-110)
	in both model performance and training cost,
	indicating that KD is not necessarily superior than SRDL in
	enhancing small model generalisation (teacher dependent). This may be partly due to the overfitting of a stronger teacher model (e.g. ResNet-110)
	which leads to less extra supervision information.
	To test this, we calculated the average cross entropy loss of the final epoch. 
	We observed 0.0087 (ResNet-110) vs 0.1637 (ResNet-32),
	which is consistent with our hypothesis.

\end{enumerate}


\begin{table} [h]
	\setlength{\tabcolsep}{0.53cm} 
	\begin{center}
		\begin{tabular}{l||c|c||c|c}
			\hline
			Query Type& \multicolumn{2}{c||}{Single-Query}&\multicolumn{2}{c}{Multi-Query}\\
			\hline
			Metrics (\%)&Rank-1&mAP&Rank-1&mAP\\
			\hline
			\hline
			SCS \cite{chen2016similarity}&51.9&26.3 &-&-\\
			G-SCNN \cite{varior2016gated}& 65.8 &39.5 &76.0& 48.4\\
			HPN \cite{liu2017hydraplus} &76.9& - &- &-\\
			MSCAN \cite{li2017learning} & 80.3 &57.5 &86.8 &66.7\\
			JLML \cite{li2017person} & 85.1 &65.5 &89.7 &74.5\\
			SVDNet \cite{sun2017svdnet} & 82.3& 62.1 &- &-\\
			PDC \cite{su2017pose} & 84.1 & 63.4 & - & - \\
			TriNet \cite{hermans2017defense} & 84.9 & 69.1 & 90.5 & 76.4 \\
			IDEAL \cite{lan2017deep}&86.7&67.5&91.3&76.2\\
			DPFL \cite{chen2018person}& 88.6 & 72.6 & 92.2 & 80.4\\
			\color{black}BraidNet-CS+SRL \cite{wang2018person}&83.7& 69.5&-&-\\
			\color{black}DaRe \cite{wang2018resource}&86.4 &69.3&- &-\\
			\color{black}MLFN \cite{chang2018multi}&90.0&\bf\color{blue} 74.3 &92.3 &\bf\color{blue}82.4\\
			\hline \hline
			ResNet-50+vanilla&87.5&69.9&91.4&78.5\\
			ResNet-50+\bf SRDL&\bf89.3&\bf73.5&\bf93.1& \bf81.5\\
			\hline
			Gain (SRDL-vanilla)
			& +1.8 & +3.6 & +1.7 & +3.0 \\
			\hline \hline
			DenseNet-121+vanilla&\bf\color{blue}90.1&74.0&\bf\color{blue}93.6&81.7\\
			DenseNet-121+\bf SRDL&\bf{\color{red}91.7}&\bf{\color{red}76.8}&\bf{\color{red} 94.2}&\bf{\color{red}83.5}\\
			\hline
			Gain (SRDL-vanilla)
			& +1.6 & +2.8 & +0.6 & +1.8 \\
			\hline
		\end{tabular}
	\end{center}
	
	\vskip -0.32cm
	\caption{Evaluation of person re-id (instance recognition) on Market-1501.
		The first/second best results are in {\bf {\color{red} red}/{\color{blue} blue}}.
	}
	\label{tab:reid}
\end{table}

\subsection{Evaluation on Person Instance Recognition}

In person re-identification (re-id) experiment,
we compared SRDL with the vanilla model learning strategy using the same CNN nets,
and also compared with ten recent the state-of-the-art re-id methods.
Two different networks are tested:
ResNet-50 (25.1M parameters) and DenseNet-121 (7.7M parameters).
Table \ref{tab:reid} shows that:
\begin{enumerate}
	\item All CNN models benefit from SRDL on the person re-id task, 
	boosting the re-id performance for both single-query and multi-query settings.

	\item SRDL trained CNNs show superior re-id performance over most state-of-the-art methods.
	In particular, SRDL trained DenseNet-121 achieves the best re-id matching rates
	among all the competitors.
\end{enumerate}

Note that, this performance gain is
obtained from a general-purpose network 
without applying any specialised person re-id model training
``bells and whistles''. 
This is in strong contrast to existing deep re-id methods \cite{varior2016gated,su2017pose,li2017learning,liu2017hydraplus}
where specially designed network architectures with complex
training process are required in order to achieve the reported results.


\subsection{Component Analysis and Discussion}
\label{sec:exp_analysis}
We further conducted SRDL component analysis
using ResNet-32 on CIFAR100.

\noindent {\bf Stage-Complete Schedule. }
Table \ref{tab:LR_schedule} compares
our stage-complete learning rate decay schedule with 
the conventional {\em stage-incomplete} counterpart.
It is evident that without the proposed schedule, 
self-referenced learning can be highly misleading
due to unreliable knowledge extracted from
the ``half-trained'' model.
This validates the aforementioned model optimisation behaviour consideration
(see the discussion underneath Eq \eqref{eq:pass_decay}).

\begin{table} [h]
	\setlength{\tabcolsep}{1.0cm} 
	\begin{tabular}{c|c}
		\hline
		Decay Strategy & Accuracy (\%) \\
		\hline
		\hline
		Stage-Incomplete &  58.11   \\ \hline
		\bf Stage-Complete & \bf 71.63  \\ 
		\hline
	\end{tabular}
	\vskip -0.1cm
	\caption{
		\footnotesize Stage-complete schedule.}	
	\label{tab:LR_schedule}
	
\end{table}

%
%
%
%
%

\noindent {\bf Random Model Restart. } 
Table \ref{tab:restart} shows that
model random restart for the second stage training in SRDL 
brings 1.90\% (71.63\%-69.73\%) accuracy gain.
{\color{black}This verifies our design motivation that 
	the discriminative knowledge is well preserved 
	in the training data and self-discovered correlation;
	Hence, random model initialisation for the second stage training of SRDL
	enables to break the optimisation search space constraint
	without losing the available information,
	and eventually improving the model generalisation capability. 
}

\begin{table} [h]
	\setlength{\tabcolsep}{1.0cm} 
	\begin{tabular}{c||c}
		\hline
		Random Restart & Accuracy  (\%) 
		\\
		\hline
		\hline
		\xmark & 69.73 
		\\ \hline
		\cmark&\bf 71.63 
		\\ 
		\hline
	\end{tabular}
	\caption{\footnotesize Random model restart.}	
	\label{tab:restart}
\end{table}

\noindent {\bf Model Ensemble.} 
Table \ref{tab:ensemble} shows that
the ensemble of ``half-trained'' and final models 
can further boost the performance by 
0.70\% (72.33\%-71.63\%) with more (double) deployment cost.
{\color{black}This suggests that the two models induced sequentially during training are 
	partially complementary, which gives rise to model ensembling diversity 
	and results in model performance boost. 
Besides, we also 
tested an ensemble of two randomly initialised networks each trained by the vanilla learning strategy for $M/2$ epochs,
obtaining the Top-1 rate
$72.02\%$ vs $72.33\%$ by SRDL. This shows that our SRDL ensemble outperforms the vanilla counterpart.

\begin{table} [h]
	\setlength{\tabcolsep}{1.0cm} 
	\centering
	\begin{tabular}{c||c}
		\hline
		Model Ensemble & Accuracy (\%) \\	
		\hline\hline
		\xmark & 71.63 \\
		\hline
		\cmark & \bf{72.33} \\
		\hline
	\end{tabular}
	\caption{\footnotesize Model ensemble.}
	\label{tab:ensemble}
\end{table}

\noindent {\bf Model Generalisation Analysis. }
As shown in \cite{keskar2016large}, model generalisation is concerned with the width of a local optimum. 
We thus examined the solutions $\bm \theta_v$ and $\bm \theta_s$ discovered by the vanilla and SRDL training algorithms, respectively. 
We added small perturbations as 
$\bm{\theta}_*(d, \bm v) = \bm \theta_* + d \cdot \bm v, \; *  \in \{v,s\}$
where $\bm v$ is a uniform distributed direction vector with a unit length, and $d \in [0,5]$ controls the change magnitude. 
The loss is quantified by the cross-entropy measurement between the predicted and ground-truth labels. Figure \ref{fig:localminima} shows the robustness of each solution against the parameter perturbation, indicating the width of local optima as $\bm \theta_v < \bm \theta_s$. This suggests that our SRDL finds a wider local minimum than the vanilla therefore more likely to generalise better.

\begin{figure}
	\centering
	\includegraphics[width=1.0\linewidth]{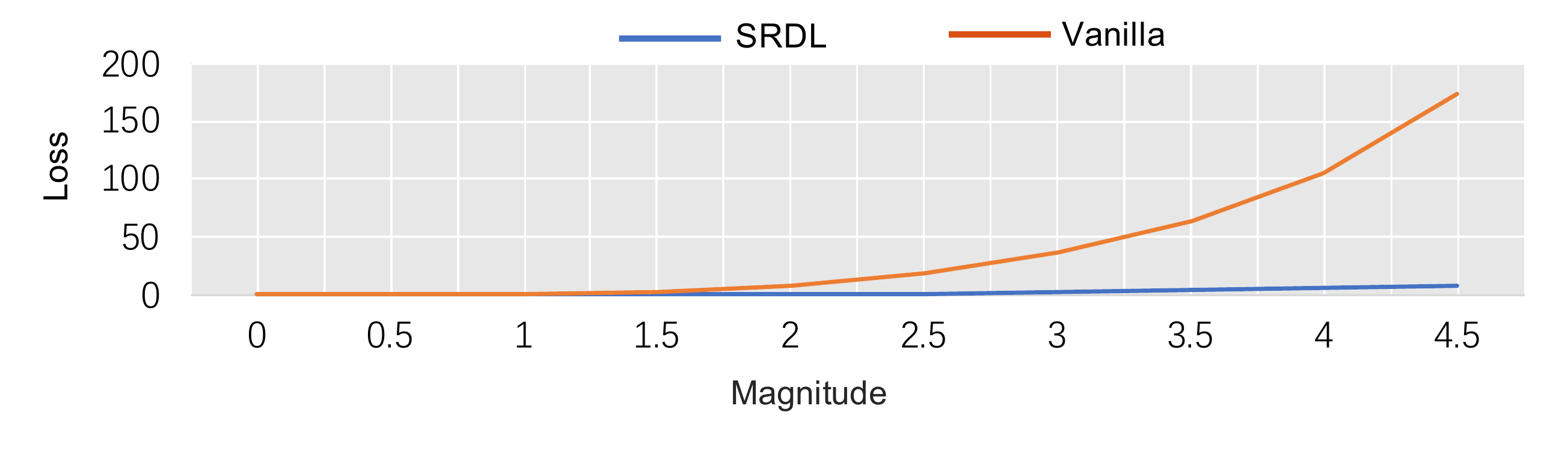} 	
	\caption{The width analysis of solution local optima.}
	
	\label{fig:localminima}
\end{figure}

\section{Conclusion}
In this work, we presented a novel Self-Referenced Deep Learning (SRDL) strategy for 
improving deep network model learning by exploiting self-discovered knowledge
in a two-stage training procedure. SRDL can train more discriminative
small and large networks with little extra computational cost.
This differs from conventional knowledge distillation which requires a
separate pre-trained large teacher model with huge 
extra computational and model training time cost.
Conceptually, SRDL is 
a principled combination of vanilla model optimisation
and existing knowledge distillation,
with an attractive trade-off
between model generalisation and model training complexity. 
Extensive experiments 
show that a variety of standard deep networks 
can all benefit from SRDL on both coarse-grained object categorisation
tasks (image classification) and fine-grained person instance
identification tasks (person re-identification).
Significantly, smaller networks benefit from more performance gains,
making SRDL specially good for low-memory and fast execution applications.
Further component analysis gives insights to the SRDL's model design considerations.

\section*{Acknowledgements}
{This work was partly supported by the China Scholarship Council, Vision Semantics Limited, the Royal Society Newton Advanced Fellowship Programme (NA150459), and Innovate UK Industrial Challenge Project on Developing and Commercialising Intelligent Video Analytics Solutions for Public Safety (98111-571149).}

\bibliographystyle{splncs04}
\bibliography{ACCV18}

\end{document}